\pdfoutput=1
\documentclass[11pt]{article}
\usepackage[utf8]{inputenc}
\usepackage[T1]{fontenc}
\usepackage{lmodern}
\usepackage[a4paper,margin=2.5cm]{geometry}
\usepackage{amsmath,amssymb}
\usepackage{graphicx}
\usepackage{booktabs}
\usepackage[table]{xcolor}
\usepackage{multirow}
\usepackage[round,authoryear]{natbib}
\usepackage[colorlinks=true,linkcolor=blue,citecolor=blue,urlcolor=blue]{hyperref}

\title{Ontological Instability and Statistical Amplification: The Paradox of ``Humanizing'' LLM-Generated Text}
\author{
  Claudiu Creanga\\
  Interdisciplinary School of Doctoral Studies\\
  University of Bucharest, Romania\\
  \texttt{claudiu.creanga@fmi.unibuc.ro}
  \and
  Liviu Dinu\\
  Faculty of Mathematics and Informatics\\
  University of Bucharest, Romania\\
  \texttt{liviu.p.dinu@unibuc.ro}
}
\date{}

\begin{document}

\maketitle

\begin{abstract}
Supervised AI-text detectors report high benchmark accuracy, but it is not clear what their decisions are based on. We analyze a RoBERTa-based detector under semantic, structural, and tokenizer-level perturbations, using the M4 dataset (N = 10,000) and controlled generations (N = 300). When Mistral-7B-Instruct was asked to make machine text sound more human, Verb Diversity rose from 0.77 to 0.92 and the outputs became easier to detect. Detection scores appear to track statistical complexity, which also leads to a 76.3\% false-positive rate on formal human writing. As a control, we evaluate event-based Latent Space detection. Paraphrasing changed 87\% of its event sequences (Jaccard = 0.067), and homoglyphs altered 70\% of the extracted verbs even though extraction still ran (Jaccard = 0.30). Its best domain AUC was 0.577. RoBERTa’s robustness seems specific to the features it uses, and structural abstraction did not make detection more robust.
\end{abstract}


\section{Introduction}

Large Language Models have made machine-text detection a practical need. Most leading systems are supervised transformer classifiers \citep{solaiman2019release, mitchell2023detectgpt}, and they report more than 99\% accuracy on standard benchmarks \citep{shishkin2024detectors}. What these classifiers actually pick up is less clear: it may be something artificial about the content, or simply statistical artifacts of particular generators.

We look at this question with RoBERTa \citep{liu2019roberta} and, as a structural point of comparison, a Latent Space detector \citep{tian2024detecting}. Both are tested with three kinds of perturbation: paraphrasing with Mistral-7B-Instruct \citep{jiang2023mistral}, changes to event sequences, and homoglyph substitution at the character level.

Our main finding is what we call \emph{Statistical Amplification}. When Mistral rewrote machine text to sound more human, lexical diversity went up, and RoBERTa’s scores usually stayed high. The detector is also poorly calibrated: at the default threshold it flagged 76.3\% of formal human texts. Homoglyph substitution evaded it in 41\% of cases, and mapping confusable characters to Latin before tokenization cut this to 2\%. In a control experiment, the Latent Space detector \citep{tian2024detecting} was affected by both paraphrasing and character-level attacks. We are not aware of earlier character-level attacks on event-based latent detectors. Taken together, the results suggest that robustness depends on the kind of features a detector uses, not on supervision as such.

\section{Related Work}

Zero-shot methods exploit probability differences between human and machine text \citep{solaiman2019release}. DetectGPT, for example, looks at the curvature of the log-probability around a text \citep{mitchell2023detectgpt}, although changes in prompting or decoding can shift the token distributions such methods depend on \citep{hans2024spotting}.

Supervised classifiers trained on large human–machine corpora are still the most common approach \citep{wang2024m4}, but they often degrade out of distribution. Part of our aim is to see whether their robustness comes from statistical complexity more than from any grasp of meaning.

Paraphrasing and synonym substitution change token distributions, and earlier work has pushed detector accuracy close to chance with them \citep{krishna2023paraphrasing, sadasivan2023can}. Our Mistral experiments point the other way: rewriting often made the signal stronger, because it raised lexical diversity.

Latent-space methods model events or discourse transitions, which are assumed to be harder for a generator to imitate \citep{tian2024detecting}. In our experiments this approach depended heavily on the domain and was of little use outside narrative text.

\section{Methodology}

\subsection{Models Under Evaluation}

We evaluate a RoBERTa classifier trained on M4 \citep{wang2024m4} and use the Latent Space approach \citep{tian2024detecting} as a structural control, approximating event triggers with verbs extracted by spaCy. If event features work as intended, they should tolerate homoglyph noise but be sensitive to paraphrasing. We also include an ensemble baseline that weights RoBERTa at 40\%, Sample Space perplexity at 10\%, and Latent Space at 50\%.

\subsection{Datasets}

We use 10,000 English samples from M4 \citep{wang2024m4}, 5,000 for validation and 5,000 for testing, balanced across WikiHow, Wikipedia, Reddit, ArXiv, and PeerRead. Most machine text comes from Bloomz; a smaller set of template-generated samples was used for baseline tests.

For the Persona Stress Test, we generated 300 additional samples using Mistral-7B-Instruct across 100 topics and three personas: Child, Neutral, and PhD.

\subsection{Metrics}

In addition to AUROC and accuracy, we use two metrics of our own. Verb Diversity (VD) is the type-token ratio of verbs:

\begin{equation}
    \text{VD} = \frac{|\text{Unique Verbs}|}{|\text{Total Verbs}|}
\end{equation}

We use it as a rough proxy for statistical complexity; low values mean that the same actions are described repeatedly. Event Sequence Preservation (ESP) is the Jaccard similarity between the verb sequences of a text before and after perturbation:

\begin{equation}
    \text{ESP} = \frac{|V_{\text{original}} \cap V_{\text{perturbed}}|}{|V_{\text{original}} \cup V_{\text{perturbed}}|}
\end{equation}

\section{Paraphrasing and Statistical Amplification}

Paraphrasing is generally expected to weaken detection, since it should remove generator-specific artifacts \citep{krishna2023paraphrasing}. With Mistral as the paraphraser, we observed the opposite.

\subsection{Iterative Paraphrasing with Mistral-7B}

We rewrote machine-generated samples iteratively with Mistral-7B-Instruct, prompting it to “humanize” the text, for example: \textit{“Rewrite the following text to sound more human, less robotic, and more varied in sentence structure.”}

\begin{table}[htbp]
\centering
\caption{Detection before and after three rounds of Mistral-7B paraphrasing.}
\label{tab:adversarial}
\begin{tabular}{lcc}
\toprule
\textbf{Metric} & \textbf{Before} & \textbf{After} \\
\midrule
Detection Score (mean) & 0.9999 & 0.9491 \\
Success Rate (<0.5) & 0\% & 5\% \\
Verb Diversity & 0.766 & 0.924 \\
Event Preservation & --- & 13.1\% \\
\bottomrule
\end{tabular}
\end{table}

After three rounds of rewriting, RoBERTa still detected 95\% of the samples. Verb Diversity rose from 0.766 to 0.924, an increase of 20.6\% (Table~\ref{tab:adversarial}), which moved the texts further from the human baseline of 0.572. By this measure, the rewritten texts were less like human writing than the originals.

\subsection{Verb Diversity by Text Type}

Table~\ref{tab:diversity} compares these values across text types.

\begin{table}[htbp]
\centering
\caption{Verb Diversity, verb count, and detection score by text type.}
\label{tab:diversity}
\begin{tabular}{lccc}
\toprule
\textbf{Text Type} & \textbf{Verb Diversity} & \textbf{Verb Count} & \textbf{Detection Score} \\
\midrule
Human (M4) & 0.572 $\pm$ 0.15 & 61.6 & 0.762 (FPR: 76.3\%) \\
Machine (Bloomz) & 0.762 $\pm$ 0.12 & 50.2 & 0.999 \\
Mistral Rewrite & 0.924 $\pm$ 0.08 & 18.1 & 0.949 \\
\bottomrule
\end{tabular}
\end{table}

\begin{figure}[htbp]
\centering
\includegraphics[width=1\textwidth]{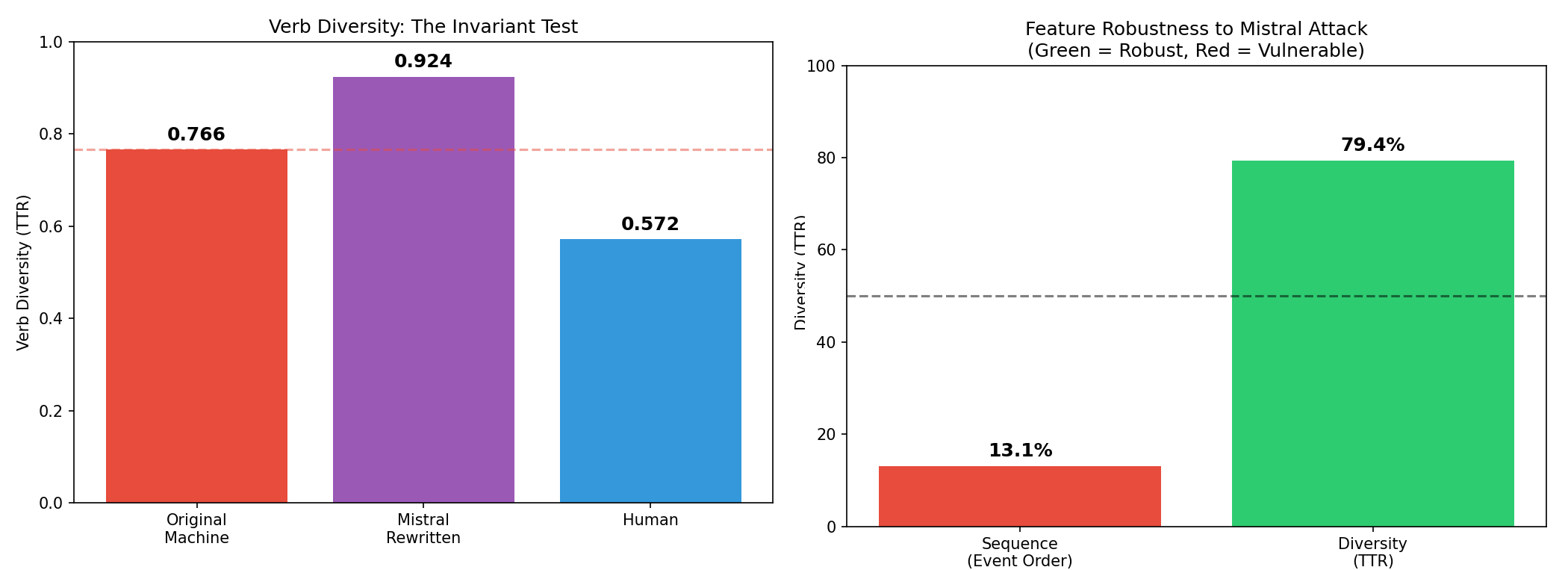}
\caption{Left: Verb Diversity before (0.766) and after (0.924) Mistral-7B rewriting, with the human baseline (0.572). Right: stability of statistical features (79.4\%) and of event sequences (13.1\%) under rewriting.}
\label{fig:invariant}
\end{figure}

Instruction-tuned models seem to have difficulty reproducing the repetition that is normal in human writing. Even when asked to “sound more human”, Mistral produced text that was more diverse than the original machine output (Figure~\ref{fig:invariant}).

\subsection{Persona Stress Test}

To look at this more directly, we generated 300 texts with Mistral-7B under three personas that ask for different levels of complexity: Child (“Explain to a 5-year-old”), Neutral, and PhD.

\begin{figure}[htbp]
\centering
\includegraphics[width=1\textwidth]{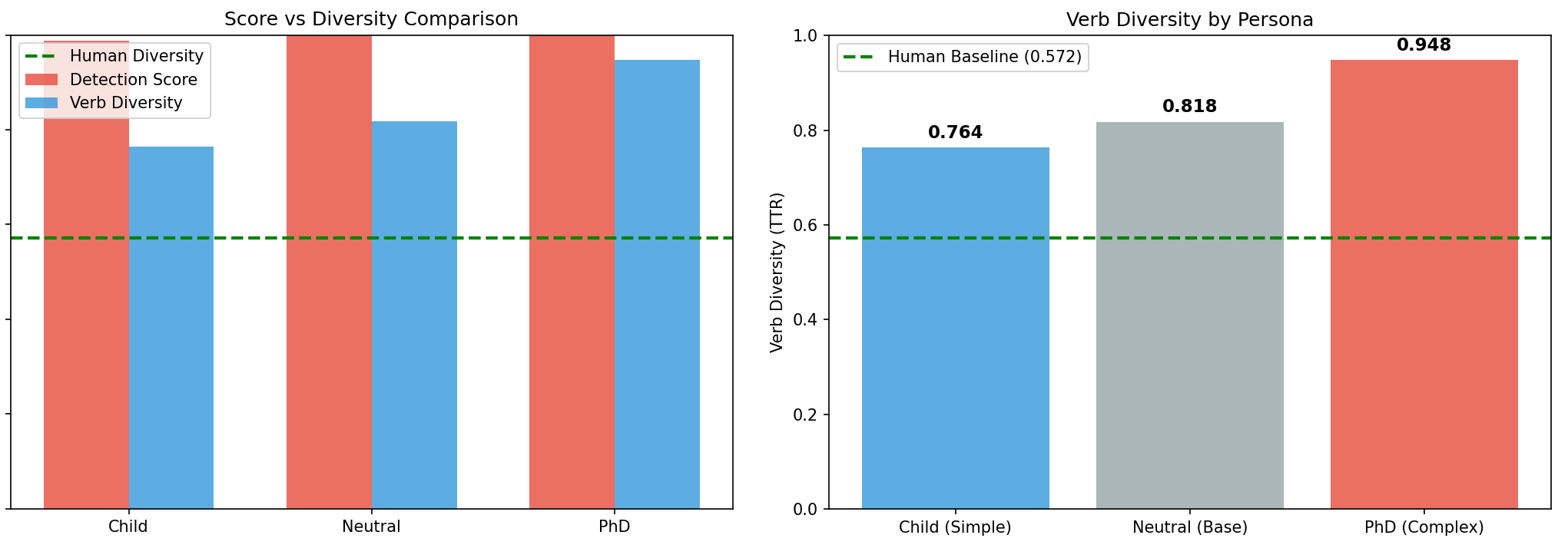}
\caption{Left: Verb Diversity against detection score for the persona samples. Right: mean Verb Diversity per persona; Child-persona texts (0.764) stay well above the human baseline (0.572), and 98.8\% of them are detected.}
\label{fig:persona}
\end{figure}

\begin{table}[htbp]
\centering
\caption{Persona Stress Test Results ($N=300$, 100 topics $\times$ 3 personas).}
\label{tab:persona}
\begin{tabular}{lccccc}
\toprule
\textbf{Persona} & \textbf{Score} & \textbf{Verb Div.} & \textbf{Tokens} & \textbf{Evasion Rate} \\
\midrule
Child (Simple) & 0.989 $\pm$ 0.10 & 0.764 $\pm$ 0.14 & 112 & 1.2\% \\
Neutral (Base) & 1.000 $\pm$ 0.00 & 0.818 $\pm$ 0.10 & 197 & 0\% \\
PhD (Complex) & 1.000 $\pm$ 0.00 & 0.948 $\pm$ 0.06 & 184 & 0\% \\
\midrule
Human Baseline & --- & 0.572 & --- & --- \\
\bottomrule
\end{tabular}
\end{table}

As Figure~\ref{fig:persona} and Table~\ref{tab:persona} show, all three personas were detected at high rates. Child-persona texts had a mean Verb Diversity of 0.764 and were detected 98.8\% of the time, while PhD texts reached 0.948 with a mean score of 1.000. The difference between human and Child-persona diversity was large ($t=-11.81$, $p < 10^{-24}$, Cohen’s $d=1.99$). The only samples that evaded detection had a diversity below 0.50.

\subsection{Compression Ratio}

We also checked whether the higher diversity shows up beyond verbs, using the Gzip compression ratio ($R = Size_{compressed}/Size_{original}$). Higher values mean the text compresses less, i.e. it has higher character-level entropy. Human text had a ratio of $0.541 \pm 0.03$ and machine text $0.587 \pm 0.04$ (Table~\ref{tab:entropy}).

Machine text was thus 8.5\% less compressible, which agrees with the Verb Diversity results. This seems to conflict with the usual view that machine text has low perplexity; we return to this in Section~\ref{sec:entropy}.

\begin{figure}[htbp]
\centering
\includegraphics[width=1\textwidth]{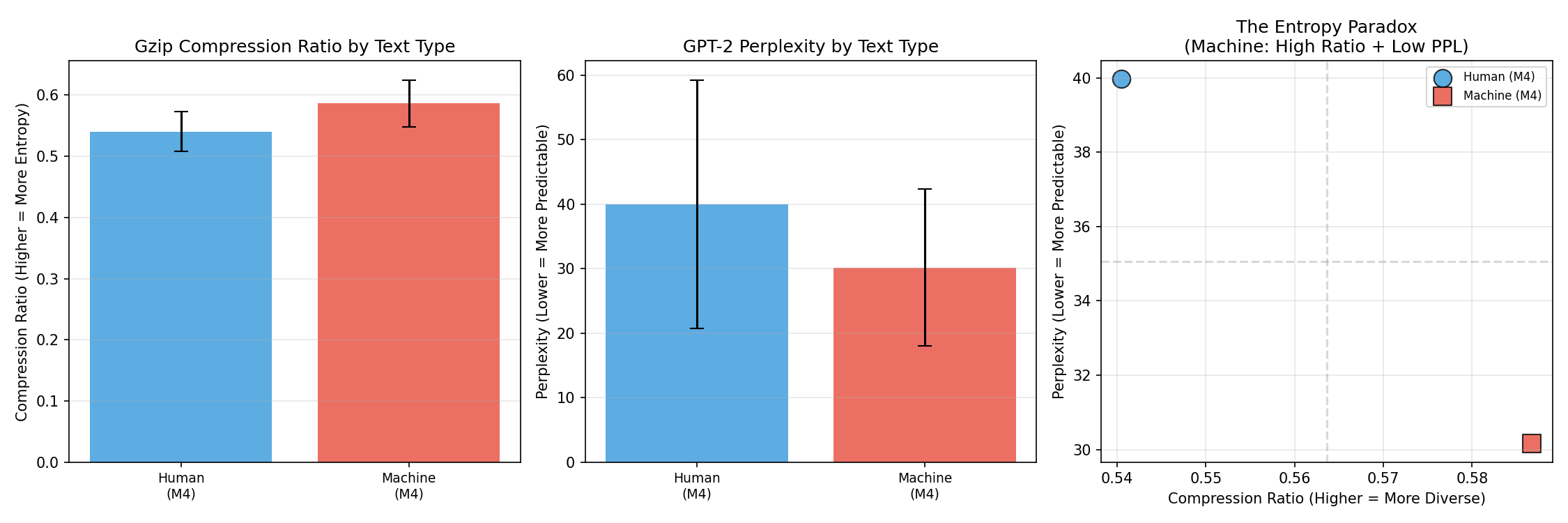}
\caption{Left and center: Gzip ratio (machine 0.587, human 0.541) and GPT-2 perplexity (machine 30.2, human 40.0). Right: per-sample scatter of the two measures.}
\label{fig:entropy}
\end{figure}

One Child-persona sample, “The Food Chain,” evaded detection completely. It repeats the same pattern throughout: “The grass gets eaten by the bunny. The bunny gets eaten by the fox.” Its Verb Diversity was 0.462 and its score 0.000. In our data, evasion required this kind of plain repetition.

\section{Tokenizer-Level Attacks}

RoBERTa handled paraphrasing well, but, as \citet{wolff2020attacking} showed earlier, it remains sensitive to character-level noise.

\subsection{Homoglyph Substitution}

Following the methodology of \citet{wolff2020attacking}, we replaced characters with visually identical Unicode homoglyphs (e.g., Cyrillic 'a' [U+0430] vs. Latin 'a' [U+0061]) at varying substitution rates.

\begin{table}[htbp]
\centering
\caption{Homoglyph Attack Results (RoBERTa).}
\label{tab:homoglyph}
\begin{tabular}{ccccc}
\toprule
\textbf{Rate} & \textbf{Success Rate} & \textbf{Mean Score} & \textbf{Score Drop} & \textbf{Avg Swaps} \\
\midrule
0\% (baseline) & 0\% & 0.981 & --- & 0 \\
5\% & 13\% & 0.868 & 0.113 & 14.6 \\
10\% & 32\% & 0.682 & 0.299 & 29.7 \\
15\% & 40\% & 0.600 & 0.380 & 45.1 \\
20\% & 41\% & 0.590 & 0.390 & 60.2 \\
30\% & 41\% & 0.584 & 0.396 & 89.3 \\
\bottomrule
\end{tabular}
\end{table}

\begin{figure}[htbp]
\centering
\includegraphics[width=1\textwidth]{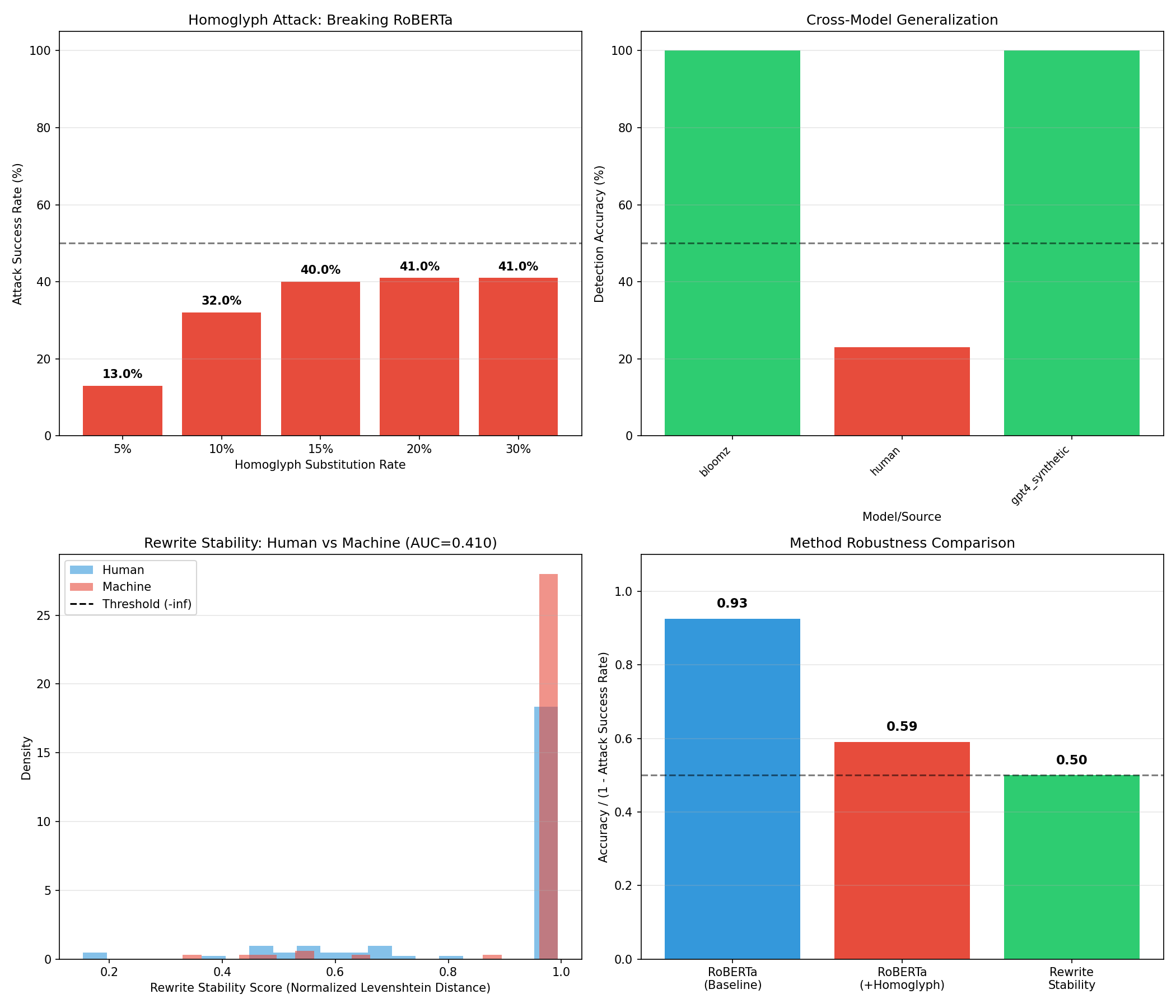}
\caption{Homoglyph evasion rate by substitution rate. Evasion reaches 13\% at 5\% substitution and levels off at about 41\% from 20\% onward.}
\label{fig:tokenizer}
\end{figure}

At 20\% substitution, the detection rate fell from 100\% to 59\%, so 41\% of samples evaded detection (Table~\ref{tab:homoglyph}). Beyond 20\%, the effect levels off. Even at 5\% (about 14 characters per text), 13\% of samples evaded (Figure~\ref{fig:tokenizer}). This is consistent with a detector that works on sub-word token IDs: homoglyphs turn common tokens into rare or unknown ones (\texttt{[UNK]}), which its learned statistics do not cover.

\subsection{Control Experiment: Latent Space Detection}

As a control, we tested whether event-based Latent Space detection behaves the other way round, resisting homoglyph noise but not paraphrasing. \citet{tian2024detecting} studied semantic perturbations; we did not find prior work applying character-level attacks to this approach.

\begin{table}[htbp]
\centering
\caption{Latent Space Vulnerability to Attacks.}
\label{tab:latent_vuln}
\begin{tabular}{lcc}
\toprule
\textbf{Attack Type} & \textbf{Jaccard Similarity} & \textbf{Extraction Failure} \\
\midrule
Homoglyph (5\%) & 0.772 & 1.4\% \\
Homoglyph (10\%) & 0.630 & 2.3\% \\
Homoglyph (30\%) & 0.299 & 5.2\% \\
\midrule
Semantic (Mistral) & 0.067 & --- \\
\bottomrule
\end{tabular}
\end{table}

Table~\ref{tab:latent_vuln} shows that the detector was vulnerable to both attacks. At 30\% homoglyph substitution, verb extraction usually still ran, but 70\% of the extracted events changed (Jaccard 0.30), and nothing in the pipeline signals the error. Mistral rewriting did even more damage (Jaccard 0.067); in one example, [analyze, discover] became [examine, reveal].

The detector needs both stable events and correct extraction. Paraphrasing breaks the first, and homoglyphs break the second.

Together with the RoBERTa results, this supports the view that robustness comes from the features a detector uses. A more abstract representation did not, by itself, make detection more robust.

\subsection{Performance by Domain}

We evaluated detection performance across five domains (Table~\ref{tab:domain}).

\begin{table}[htbp]
\centering
\caption{Domain-Specific Performance (AUC).}
\label{tab:domain}
\begin{tabular}{lccc}
\toprule
\textbf{Domain} & \textbf{RoBERTa} & \textbf{Sample Space} & \textbf{Latent Space} \\
\midrule
WikiHow & 1.000 & 0.559 & \textbf{0.577} \\
Wikipedia & 1.000 & 0.497 & 0.413 \\
Reddit & 1.000 & 0.606 & 0.459 \\
ArXiv & 1.000 & 0.755 & 0.385 \\
PeerRead & 1.000 & 0.730 & 0.393 \\
\bottomrule
\end{tabular}
\end{table}

The Latent Space detector was above chance only on WikiHow (AUC 0.577), and it scored below chance on ArXiv (0.385) and PeerRead (0.393).

We see three reasons for this:

\begin{itemize}
\item Technical writing relies on nominalizations (“analysis” instead of “analyze”), so the extracted events are unreliable.
\item Human and machine academic abstracts both contain about 50–60 verbs, with similar distributions, so there is little signal to separate them.
\item Paraphrasing changed 87\% of event sequences, so events behave more like wording choices than like stable properties of the content.
\end{itemize}
Even on narrative text the gain is small, and on technical text the detector does worse than chance. Table~\ref{tab:attack_surface} compares the two detectors.

\begin{table}[htbp]
\centering
\caption{Attack results and best-domain AUC for the two detectors.}
\begin{tabular}{lccc}
\toprule
\textbf{Method} & \textbf{Semantic Attack} & \textbf{Tokenizer Attack} & \textbf{Best Domain AUC} \\
\midrule
RoBERTa & \textbf{Robust (95\%)} & Vulnerable (59\%) & 1.000 (all) \\
Latent Space & \textbf{Poor (13\%)} & Vulnerable (30\%) & 0.577 (WikiHow) \\
\bottomrule
\label{tab:attack_surface}
\end{tabular}
\end{table}

RoBERTa resists paraphrasing but not homoglyphs, whereas the Latent Space detector is affected by both.

\subsection{Lexical Entropy and Perplexity}
\label{sec:entropy}

Machine text is often described as low-entropy \citep{mitchell2023detectgpt}. On 200 M4 samples, we measured two different things: lexical entropy, through the Gzip ratio, and predictability to a language model, through GPT-2 perplexity.

\begin{table}[htbp]
\centering
\caption{Gzip compression ratio and GPT-2 perplexity (N = 200).}
\label{tab:entropy}
\begin{tabular}{lcc}
\toprule
\textbf{Text Type} & \textbf{Gzip Ratio} & \textbf{Perplexity (GPT-2)} \\
\midrule
Human & 0.541 $\pm$ 0.03 & 40.0 $\pm$ 19.3 \\
Machine & 0.587 $\pm$ 0.04 & 30.2 $\pm$ 12.2 \\
\bottomrule
\end{tabular}
\end{table}

Machine text was more predictable ($PPL_{machine}=30.2$ vs. $PPL_{human}=40.0$) but less compressible ($Ratio_{machine}=0.587$ vs. $Ratio_{human}=0.541$). A plausible reading is that LLMs choose safe, high-probability continuations while drawing on a wider vocabulary than human writers do. RoBERTa may be picking up this lexical density (Figure~\ref{fig:entropy}).

\subsection{Alternative Detection Methods}

We also tried two other detection methods. The Binoculars score, a perplexity ratio, reached only 50.3\% accuracy (AUC = 0.217). Rewrite stability (RAIDAR), which measures edit distance after T5 rewriting, did not separate the classes (AUC = 0.492), so in our setting machine text was not more “stable” under rewriting.

\section{Calibration and Defense}

\subsection{Threshold Calibration}

At the default threshold of 0.5, the detector labelled 76.3\% of human ArXiv abstracts as machine-generated. Raising the threshold to $\tau=0.9999$ removed these false positives, but machine recall dropped to 1.2\% and homoglyph attack success rose to 97\% (Table~\ref{tab:calibration}). We could not find a single threshold that worked well.

\begin{table}[htbp]
\centering
\caption{Detector behavior at the default and strict thresholds.}
\label{tab:calibration}
\begin{tabular}{lcc}
\toprule
\textbf{Metric} & \textbf{Default (0.5)} & \textbf{Strict (0.9999)} \\
\midrule
False Positive Rate (Human) & \textbf{76.3\%} & 0\% \\
True Positive Rate (Machine) & 98.5\% & 1.2\% \\
Homoglyph Attack Success & 30\% & 97\% \\
\bottomrule
\end{tabular}
\end{table}

\begin{figure}[htbp]
\centering
\includegraphics[width=1\textwidth]{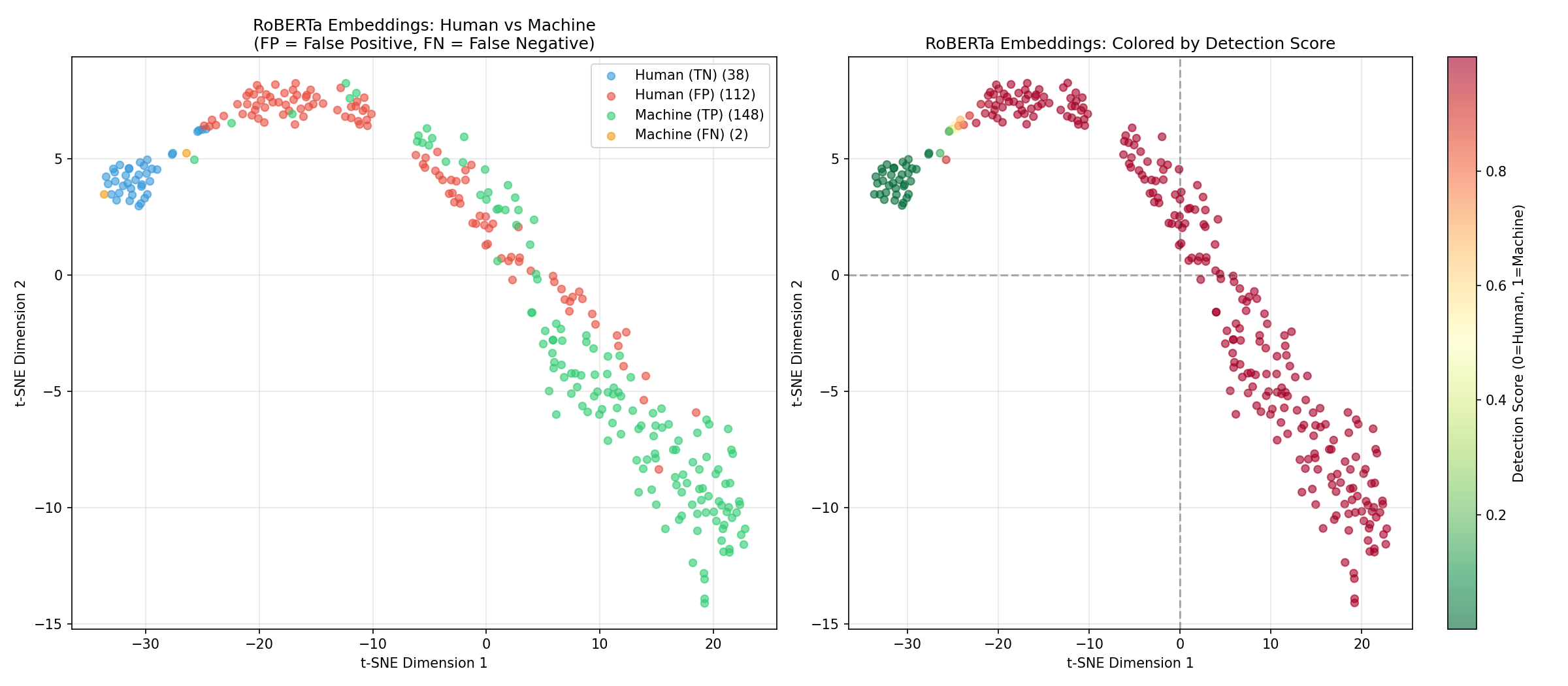}
\caption{T-SNE projection of RoBERTa embeddings. Formal human text (blue triangles) falls inside the machine cluster (red circles).}
\label{fig:tsne}
\end{figure}

In the T-SNE projection (Figure~\ref{fig:tsne}), formal human text falls inside the machine cluster, while casual Reddit posts stay apart. The detector seems to respond to formality more than to anything specific to machine text.

\subsection{Confusable-Character Sanitization}

NFKC normalization does not map look-alike characters from other scripts to Latin, so it had no effect on the attack. We therefore added a lookup table of common confusables, such as Cyrillic ‘c’ (U+0441) and ‘e’ (U+0435), and replaced them with their Latin counterparts before tokenization.

\begin{table}[htbp]
\centering
\caption{Homoglyph attack success before and after sanitization.}
\label{tab:sanitization}
\begin{tabular}{lc}
\toprule
\textbf{Condition} & \textbf{Attack Success Rate} \\
\midrule
Original (no attack) & 0\% \\
Homoglyph Attack (20\%) & 41\% \\
After NFKC Normalization & 41\% (no change) \\
After confusable mapping & \textbf{2\%} \\
\bottomrule
\end{tabular}
\end{table}

\begin{figure}[htbp]
\centering
\includegraphics[width=1\textwidth]{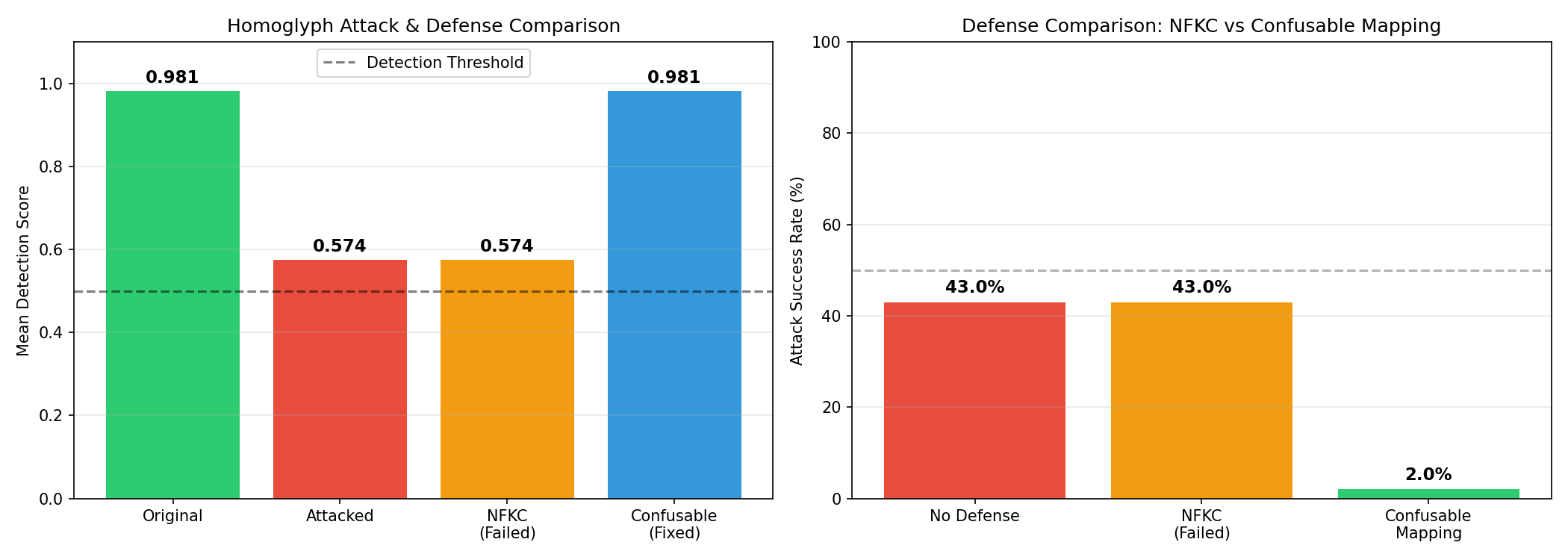}
\caption{Homoglyph attack success before (41\%) and after (2\%) confusable mapping.}
\label{fig:aggressive_sanitization}
\end{figure}

With this mapping, attack success fell from 41\% to 2\% (Table~\ref{tab:sanitization}; Figure~\ref{fig:aggressive_sanitization}). For a deployed detector, this kind of preprocessing seems necessary.

\section{Other Attacks}

\subsection{Syntax Simplification}

Simplification with Flan-T5-Large \citep{chung2022flan} led to evasion in 6\% of samples, with a mean score drop of 0.031. One sample, however, dropped by 0.997. As with “The Food Chain”, the few successful cases were texts that had become much simpler.

\subsection{Recursive Back-Translation}

\begin{figure}[htbp]
\centering
\includegraphics[width=1\textwidth]{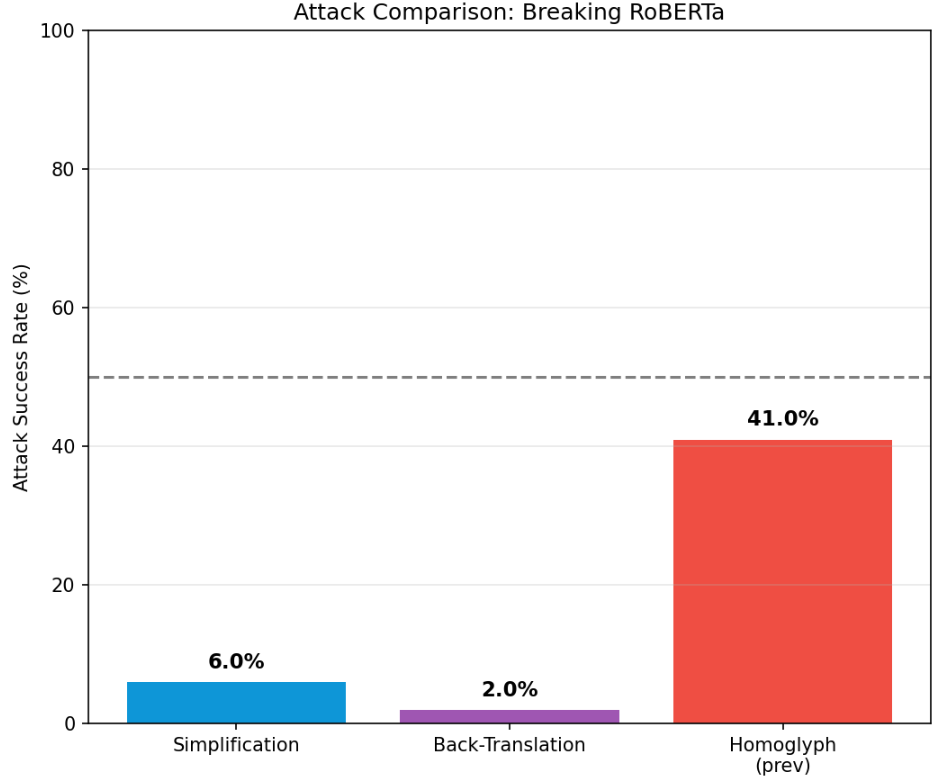}
\caption{Evasion rates for simplification (6\%), back-translation (2\%), and homoglyph substitution (41\%). The dashed line marks 50\%.}
\label{fig:sanitization}
\end{figure}

Round-trip translation (EN$\rightarrow$FR$\rightarrow$DE$\rightarrow$EN) led to evasion in 2\% of samples, with a mean score drop of 0.018, so the detection signal largely survived translation (Figure~\ref{fig:sanitization}).

\subsection{Greedy Word-Swap Attack}

We also ran a black-box greedy word-swap attack, using BERT-base \citep{devlin2019bert} to propose substitutions and keeping, at each step, the one that lowered RoBERTa’s score the most.

On 49 of the 50 samples, no substitution lowered the score enough to be worth making. If the detector relied on a few giveaway words, the search should have found them; the result suggests that the signal is spread across the whole text.

\section{Discussion}

\subsection{Lexical Diversity as the Main Signal}

RoBERTa’s scores are closely associated with lexical diversity, although correlated syntactic or token-level features may also play a role. This association accounts for the three main observations: paraphrasing does not help an attacker, homoglyphs do, and formal human prose is often misclassified.

\subsection{Feature Choice and Robustness}

The Latent Space control suggests that robustness depends on which features a detector uses. Statistical and structural features failed under different perturbations.

Working at the level of events did not make the detector more robust. It added new ways to fail, such as extraction errors, while the sensitivity to paraphrasing remained.

For an attacker, paraphrasing appears counterproductive. In our experiments, evasion required lowering complexity through repetition, attacking the tokenizer with homoglyphs, or bringing Verb Diversity below 0.50.

In practice, a deployed detector should at least sanitize confusable characters, calibrate thresholds per domain, and combine statistical with structural signals.

\section{Conclusion and Future Work}

This paper examined why a supervised AI-text detector is robust to some attacks and not to others. The main findings are:

\begin{enumerate}
\item Paraphrasing with Mistral increased lexical diversity and made texts more detectable (Statistical Amplification).
\item Homoglyph substitution evaded the detector in 41\% of cases, in line with earlier work.
\item The detector confused formality with machine generation, with a 76\% false-positive rate on academic text.
\item Mapping confusable characters to Latin reduced homoglyph attack success to 2\%.
\end{enumerate}
Overall, the detector behaves more like a filter for statistical complexity than like a system that recognizes what makes a text machine-written.

Future work should check whether Statistical Amplification holds for newer generators, other languages, and other architectures, and should look for calibration methods that can tell academic formality apart from machine generation.

\section{Limitations}

Our main experiments use a single RoBERTa detector and English text only, so we cannot say how the results transfer to other architectures or languages. The Latent Space detector uses spaCy verbs as a stand-in for events, and a full event extractor might behave differently. Homoglyphs are also easy to detect with Unicode analysis; our claim is only that detectors without sanitization miss them.

\section{Ethics Statement}

This study reports detector vulnerabilities to support more reliable systems. It uses the public M4 dataset and open-weight models, collects no private data, and involves no human participants.

\paragraph{Generative AI Declaration.} The authors used AI tools for writing code, language polishing, and reference management. All outputs were reviewed and edited by the authors, who take full responsibility for the content of this article.


\begin{thebibliography}{13}

\bibitem[Chung et al.(2022)]{chung2022flan}
Chung, H. W., L. Hou, S. Longpre, B. Zoph, Y. Tay, W. Fedus, Y. Li, et al. 2022. “Scaling Instruction-Finetuned Language Models.” \textit{arXiv preprint arXiv:2210.11416}. \url{https://arxiv.org/abs/2210.11416}

\bibitem[Devlin et al.(2019)]{devlin2019bert}
Devlin, J., M.-W. Chang, K. Lee, and K. Toutanova. 2019. “BERT: Pre-training of Deep Bidirectional Transformers for Language Understanding.” In \textit{Proceedings of the 2019 Conference of the North American Chapter of the Association for Computational Linguistics: Human Language Technologies, Volume 1}, 4171–4186. Association for Computational Linguistics. \url{https://aclanthology.org/N19-1423/}

\bibitem[Gritsai et al.(2024)]{shishkin2024detectors}
Gritsai, G., A. Voznyuk, A. Grabovoy, and Y. Chekhovich. 2024. “Are AI Detectors Good Enough? A Survey on Quality of Datasets with Machine-Generated Texts.” \textit{arXiv preprint arXiv:2410.14677}. \url{https://arxiv.org/abs/2410.14677}

\bibitem[Hans et al.(2024)]{hans2024spotting}
Hans, A., A. Schwarzschild, V. Cherepanova, H. Kazemi, A. Saha, M. Goldblum, J. Geiping, and T. Goldstein. 2024. “Spotting LLMs with Binoculars: Zero-Shot Detection of Machine-Generated Text.” In \textit{Proceedings of the 41st International Conference on Machine Learning}, 17519–17537. PMLR 235. \url{https://proceedings.mlr.press/v235/hans24a.html}

\bibitem[Jiang et al.(2023)]{jiang2023mistral}
Jiang, A. Q., A. Sablayrolles, A. Mensch, C. Bamford, D. S. Chaplot, D. de las Casas, et al. 2023. “Mistral 7B.” \textit{arXiv preprint arXiv:2310.06825}. \url{https://arxiv.org/abs/2310.06825}

\bibitem[Krishna et al.(2023)]{krishna2023paraphrasing}
Krishna, K., Y. Song, M. Karpinska, J. Wieting, and M. Iyyer. 2023. “Paraphrasing Evades Detectors of AI-Generated Text, but Retrieval Is an Effective Defense.” In \textit{Advances in Neural Information Processing Systems} 36: 27469–27500. \url{https://proceedings.neurips.cc/paper_files/paper/2023/hash/575c450013d0e99e4b0ecf82bd1afaa4-Abstract-Conference.html}

\bibitem[Liu et al.(2019)]{liu2019roberta}
Liu, Y., M. Ott, N. Goyal, J. Du, M. Joshi, D. Chen, O. Levy, M. Lewis, L. Zettlemoyer, and V. Stoyanov. 2019. “RoBERTa: A Robustly Optimized BERT Pretraining Approach.” \textit{arXiv preprint arXiv:1907.11692}. \url{https://arxiv.org/abs/1907.11692}

\bibitem[Mitchell et al.(2023)]{mitchell2023detectgpt}
Mitchell, E., Y. Lee, A. Khazatsky, C. D. Manning, and C. Finn. 2023. “DetectGPT: Zero-Shot Machine-Generated Text Detection Using Probability Curvature.” In \textit{Proceedings of the 40th International Conference on Machine Learning}, 24950–24962. PMLR. \url{https://proceedings.mlr.press/v202/mitchell23a.html}

\bibitem[Sadasivan et al.(2023)]{sadasivan2023can}
Sadasivan, V. S., A. Kumar, S. Balasubramanian, W. Wang, and S. Feizi. 2023. “Can AI-Generated Text Be Reliably Detected?” \textit{arXiv preprint arXiv:2303.11156}. \url{https://arxiv.org/abs/2303.11156}

\bibitem[Solaiman et al.(2019)]{solaiman2019release}
Solaiman, I., M. Brundage, J. Clark, A. Askell, A. Herbert-Voss, J. Wu, A. Radford, D. Krueger, J. W. Kim, S. Kreps, et al. 2019. “Release Strategies and the Social Impacts of Language Models.” \textit{arXiv preprint arXiv:1908.09203}. \url{https://arxiv.org/abs/1908.09203}

\bibitem[Tian et al.(2024)]{tian2024detecting}
Tian, Y., Z. Pan, and N. Peng. 2024. “Detecting Machine-Generated Long-Form Content with Latent-Space Variables.” In \textit{Findings of the Association for Computational Linguistics: EMNLP 2024}, 10394–10408. Association for Computational Linguistics. \url{https://aclanthology.org/2024.findings-emnlp.608/}

\bibitem[Wang et al.(2024)]{wang2024m4}
Wang, Y., J. Mansurov, P. Ivanov, J. Su, A. Shelmanov, A. Tsvigun, C. Whitehouse, et al. 2024. “M4: Multi-Generator, Multi-Domain, and Multi-Lingual Black-Box Machine-Generated Text Detection.” In \textit{Proceedings of the 18th Conference of the European Chapter of the Association for Computational Linguistics (Volume 1: Long Papers)}, 1369–1407. Association for Computational Linguistics. \url{https://aclanthology.org/2024.eacl-long.83/}

\bibitem[Wolff and Wolff(2020)]{wolff2020attacking}
Wolff, M., and S. Wolff. 2020. “Attacking Neural Text Detectors.” \textit{arXiv preprint arXiv:2002.11768}. \url{https://arxiv.org/abs/2002.11768}

\end{thebibliography}
\end{document}